\documentclass[lettersize,journal]{IEEEtran}
\usepackage{amsmath,amsfonts}
\usepackage{algorithmic}
\usepackage{algorithm}
\usepackage{array}
\usepackage[caption=false,font=normalsize,labelfont=sf,textfont=sf]{subfig}
\usepackage{textcomp}
\usepackage{stfloats}
\usepackage{url}
\usepackage{verbatim}
\usepackage{graphicx}
\usepackage{cite}
\usepackage[T1]{fontenc} 
\usepackage{amsmath}
\usepackage{amsthm}
\usepackage{amssymb}
\usepackage{setspace}
\usepackage{enumitem}

\begin{document}

\title{Multiple-gain Estimation for Running Time of Evolutionary Combinatorial Optimization}
\author{Min Huang, Pengxiang Chen, Han Huang*, Tongli He, Yushan Zhang, Zhifeng Hao}
\maketitle
\begin{abstract}
	The running-time analysis of evolutionary combinatorial optimization is a fundamental topic in evolutionary computation. Its current research mainly focuses on specific algorithms for simplified problems due to the challenge posed by fluctuating fitness values. This paper proposes a multiple-gain model to estimate the fitness trend of population during iterations. The proposed model is an improved version of the average gain model, which is the approach to estimating the running time of evolutionary algorithms for numerical optimization. The improvement yields novel results of evolutionary combinatorial optimization, including a briefer proof for the time complexity upper bound in the case of (1+1) EA for the Onemax problem, two tighter time complexity upper bounds than the known results in the case of (1+$\lambda$) EA for the knapsack problem with favorably correlated weights and a closed-form expression of time complexity upper bound in the case of (1+$\lambda$) EA for general $k$-MAX-SAT problems. The results indicate that the practical running time aligns with the theoretical results, verifying that the multiple-gain model is more general for running-time analysis of evolutionary combinatorial optimization than state-of-the-art methods.
\end{abstract}


\section{Introduction}
\IEEEPARstart{R}{unning-time} analysis of evolutionary algorithms (EAs) is an important and challenging task. The study of running time helps explain why EAs can solve problems quickly, and improve the efficiency of algorithms. The first hitting time (FHT) is a commonly applied concept for estimating the running time of EAs, reflecting the number of iterations required for an EA to first attend the global optimum\cite{he2016average}. In addition, the expected first hitting time (EFHT) refers to the average number of iterations that an EA needs to attend the global optimum initially\cite{chen2009new}. Therefore, the aim of running-time analysis can be described as analyzing the EFHT of EAs.

In the past two decades, great advancements have been made in running-time analysis within the field of evolutionary computation. In contrast to evolutionary numerical optimization, the
studies related to evolutionary combinatorial optimization are still
at an early stage. As a wide range of engineering tasks are combinatorial optimization problems, studying evolutionary combinatorial optimization is crucial. Consequently, this paper considers the EFHT of evolutionary combinatorial optimization.

There has been a notable increase in theoretical research
of evolutionary numerical optimization in the last decade. Recently, the average gain model has been applied to the running-time analysis of EAs used in practice for continuous optimization \cite{huang2019experimental}. The average gain model was introduced in \cite{hhan2014} for estimating the running time of (1+1)EA on the sphere function. Zhang et al. \cite{yushan2016first} combined martingale theory with the stopping time theory to extend the average gain model. Subsequently, Huang et al. \cite{huang2019experimental} introduced an experimental approach for estimating the EFHT of evolutionary numerical optimization in conjunction with the average gain model. The method utilizes surface fitting techniques to simplify analysis by replacing complex mathematical calculations. However, the surface fitting function designed in this work relies on continuity. Since the gain in combinatorial optimization problems is typically discrete, this experimental method cannot be directly applied to evolutionary combinatorial optimization. Although the average gain model has limitations for evolutionary combinatorial optimization, the concepts and experimental methods in \cite{huang2019experimental} and \cite{yushan2016first} provide the theoretical foundation for this paper.

Current research on the running-time analysis of evolutionary combinatorial optimization primarily focuses on case-specific studies. For example,  Benjamin et al. \cite{doerr2023first, zheng2023running-time} analyzed the time complexity of NSGA-II on the OneMinMax and OneJumpZeroJump functions. Doerr and Qu \cite{doerr2023runtime} proved that the NSGA-II can optimize the OneJumpZeroJump function asymptotically more efficient when crossover is applied. Lu et al. \cite{lu2024towards} further demonstrated that the running times of R-NSGA-II on the OneJumpZeroJump and OneMinMax functions are all asymptotically more efficient than the NSGA-II. Bian et al. \cite{bian2024archive,bian2023stochastic} reduced the running times of NSGA-II and SMS-EMOA on certain problems. Furthermore, they proved the time complexity of the standard NSGA-II on the LOTZ problem\cite{bian2022better}. Lai et al. \cite{lai2020analysis} analyzed the running time of multi-objective EAs for bi-objective traveling salesman problem (1,2). Qian et al. \cite{qian2024quality} demonstrated the time complexity of quality-diversity algorithm MAP-Elites on monotone approximately submodular maximization with a size constraint, and set cover. Additionally, Dang et al. \cite{dang2023analysing} analyzed the time complexity of NSGA-II on LeadingOnesTrailingZeroes under a specific noise model. Since the authors of the above studies primarily focused on specific cases, there remains a lack of results concerning general analysis tools for evolutionary combinatorial optimization.

Among the few existing general analysis tools, the drift analysis introduced by He and Yao \cite{2001Drift} is the primary approach used by researchers to analyze the running time of evolutionary combinatorial optimization. The core concept of drift analysis involves converting the expected progress of fitness values in one iteration into estimates for the EFHT\cite{doerr2021survey}. Jägersküpper \cite{MarkovChain} combined Markov chains with drift analysis, significantly improving the time complexity upper bound of (1+1)EA for linear functions. Oliveto and Witt \cite{oliveto2011simplified} proposed a simplified drift analysis theorem to establish the time complexity lower bounds of (1+1) EA for problems such as Needle, OneMax, and maximum matching. Witt \cite{witt2013tight} subsequently obtained tight time complexity bounds of (1+1) EA for linear function by applying multiplicative drift analysis. Fajardo et al. \cite{fajardo2023runtime} produced a variance drift theorem, which utilizes the variance of the process to overcome weak drift tendencies. Despite its considerable generality in analyzing evolutionary combinatorial optimization, drift analysis has certain limitations, such as computing the drift is non-trivial and depends on a right potential function\cite{doerr2021survey,doerr2019theory}. Furthermore, the fluctuating fitness values in most combinatorial optimization problems complicate the estimation of the expected progress of fitness values in a single iteration.

In summary, the running-time analysis of evolutionary combinatorial optimization primarily focuses on specific case studies. Existing general analysis tools are limited by the fluctuating fitness values in most combinatorial optimization problems. Therefore, this paper aims to overcome the difficulty of estimating the expected progress of fitness values, and propose a general approach for analyzing the running time of evolutionary combinatorial optimization.

Inspired by the average gain model\cite{yushan2016first,huang2019experimental}, this paper proposes a multiple-gain model based on the law of large numbers to estimate the average-case and worst-case upper bounds of EFHT for evolutionary combinatorial optimization. The multiple-gain model constructs the multiple-gain function to estimate the fitness trend of population during iterations. In most studies on the running-time analysis of evolutionary combinatorial optimization, the methods used are dependent on specific conditions. The proposed model provides a unified framework to estimate the running time of various evolutionary combinatorial optimization instances, without the need for specific conditions. Our contributions are listed below.
\begin{itemize} 
	\item Providing a more general running-time analysis method than the existing state-and-the-art methods for evolutionary combinatorial optimization.
	\item Addressing the difficulty of analyzing combinatorial optimization problems with the average gain model by approaching it from the perspective of multiple-gain. In fact, the average gain is a special case of the multiple-gain.
	\item Providing a briefer proof for the average-case upper bound of EFHT in the case of (1+1) EA for the Onemax problem. Previously, this instance has been extensively investigated in \cite{giessen2017interplay,oliveto2011simplified, doerr2014impact,droste1998rigorous,adak2024runtime}. This paper gives a series form average-case upper bound of EFHT that is consistent with the known asymptotic upper bound $O(nlogn)$.
	\item Providing two tighter average-case upper bounds of EFHT under certain conditions than the known results \cite{neumann2018running-time} in the case of (1+$\lambda$) EA for the knapsack problem with favorably correlated weights. Indeed, the conclusion in \cite{neumann2018running-time} represents a special case of our result when $\lambda=1$.
	\item Providing a first closed-form expression for the average-case upper bound of EFHT in the case of a class of (1+$\lambda$) EA for general $k$-MAX-SAT problems. Previously, various specific instances of the $k$-MAX-SAT problem have been investigated in \cite{buzdalov2017running-time,doerr2015improved,sutton2012parameterized}. This paper attempts to analyze the running time of EAs for the general $k$-MAX-SAT problem. 
\end{itemize}

The remainder of this article is structured as follows. Section II presents a detailed description of the multiple-gain model, along with a brief overview of the relevant background. In Section III, we utilize the multiple-gain model to estimate the average-case upper bounds of EFHT for three evolutionary combinatorial optimization instances. Section IV presents experimental results of the average-case and worst-case upper bounds of EFHT for these three instances. Section V summarizes the content of this article.

\section{Multiple-gain Model}
This section begins with an overview of fundamental concepts in single-objective combinatorial optimization, followed by the average gain model. Next, We describe the multiple-gain model.
\subsection{Preliminaries}
We consider the following optimization problem\cite{back1997evolutionary}:
\begin{center}
	$min\quad f(x)\in S$\\
	$s. t. \quad x\in \left\{0, 1\right\}^{n}$
\end{center}
where $f(x)$ maps $\left\{0, 1\right\}^{n}$ to the target space $S$, with $n$ representing the encoding length.

Let $r_{i}\in S$ denote a fitness value, and $r_{m}\in S$ denote the biggest fitness value where $0\le i \le m$. For convenience, we presume $ S=\left\{r_{0}, r_{1}, \dots, r_{m}\right\}$ where $r_{0}<r_{1}<\dots<r_{m}$. 

Let $X_{opt}\in \left\{0, 1\right\}^{n}$ denote a global optimum where $f(X_{opt})=r_{0}$, and $X_{t}\in \left\{0, 1\right\}^{n}$ denote an offspring
individual in the $t$-th generation.

The optimization process of EAs is characterized by randomness. Within the same parent population, it is possible to generate different offspring populations. Therefore, the optimization process can be regarded as a stochastic process. Let $(\Omega, F, P)$ be a probability space, and let $\left \{ Y_{t} \right \} _{t=0}^{\infty}$ represent a non-negative stochastic process defined on this space. Let $F_{t}=\sigma(Y_{0}, Y_{1}, \dots, Y_{t})$ denote the natural filtration of $F$.

In the following, we give the specific definitions of supermartingale and stopping time\cite{durrett2019}. 

\textbf{\emph{Definition 1: }}{Let $\left \{ Y_{t} \right \}_{t=0}^{\infty}$ and $\left \{ Z_{t} \right \}_{t=0}^{\infty}$ be  two stochastic processes. $\left \{ Y_{t} \right \}_{t=0}^{\infty}$ is a supermartingale relating to $\left \{ Z_{t} \right \}_{t=0}^{\infty}$ if $Y_{t}$ depends on $Z_{0},Z_{1},\dots,Z_{t}$, $E(max\left\{0, -Y_{t}\right\})<\infty$, $E(Y_{t+1}|Z_{0}, Z_{1}, \dots, Z_{t})\le Y_{t}$ for all $ t\ge 0$.}

\textbf{\emph{Definition 2: }}{Let $\left \{ Y_{t} \right \}_{t=0}^{\infty}$ be a stochastic process and $T\in\mathbb{N}_0$ be a random variable. $T$ is a stopping time relating to $\left \{ Y_{t} \right \}_{t=0}^{\infty}$ if $\left\{T\le n\right\}\in F_{t}$ for all $n=0, 1, 2, \dots$. }

The iteration at which the global optimum is first found is called the first hitting time (FHT)\cite{he2016average}. The definition of FHT is given below. 

\textbf{\emph{Definition 3: }}{Let $\left \{ Y_{t} \right \}_{t=0}^{\infty}$ denote a stochastic process, where $Y_{t}\ge 0$ holds for all $t\ge 0$. The first hitting time of EAs is defined as $T_{\varepsilon}=min\left\{t\ge 0: Y_{t}\le \varepsilon\right\}$ where $\varepsilon$ represents the target accuracy of EAs. }
Specifically, $T_{0}=min\left\{t\ge 0:Y_{t}=0\right\}$ is a stopping time relating to $\left \{ Y_{t} \right \}_{t=0}^{\infty}$ \cite{yushan2016first}.

Moreover, the expected first hitting time (EFHT) of EAs is denoted by $E(T_{\varepsilon})$, which is the expected value of FHT. The EFHT represents the iteration number on average needed for EAs to first attend a global optimum\cite{yu2008new}. 
\subsection{Average Gain Model and Multiple-Gain Model}
The average gain represents the difference between the fitness value of parent individual $X_{t}$ and the fitness value of offspring individual $X_{t+1}$. The formal definition of average gain is presented below\cite{yushan2016first}.

\textbf{\emph{Definition 4: }}{The gain at generation $t$ is}
\begin{align*}
	g_{t}=f(X_{t})-f(X_{t+1}). 
\end{align*}
Let $H_{t}=\sigma(f(X_{0}), f(X_{1}), \dots, f(X_{t}))$. The average gain at generation $t$ is 
\begin{align*}
	E(g_{t}|H_{t})=E(f(X_{t})-f(X_{t+1})|H_{t}). 
\end{align*}

Similar to quality gain, the average gain indicates the expected progress of EAs within one iteration\cite{akimoto2017quality}. A greater average gain implies a quicker convergence towards the global optimum, thereby improving the efficiency of each iteration in the optimization process.

Inspired by the average gain model, we propose the multiple-gain model for analyzing the EFHT of evolutionary combinatorial optimization. The definition of multiple-gain and expected multiple-gain are presented in Definition 5.

\textbf{\emph{Definition 5: }}{For a given $k$, the multiple-gain at generation $t$ is}
\begin{align*}
	f(X_{t})-f(X_{t+k}). 
\end{align*}
Let $H_{t}=\sigma(f(X_{0}), f(X_{1}), \dots, f(X_{t}))$. For a given $k$, the expected multiple-gain at generation $t$ is 
\begin{align*}
	G(t,k)=E(f(X_{t})-f(X_{t+k})|H_{t}). 
\end{align*}

The average gain is a special case of the expected multiple-gain model when $k=1$. The following lemma, which was proved in \cite{yushan2016first}, will be used in the proof of Theorem 1.

\textbf{\emph{Lemma 1: }}{Let $\left \{ S_{t} \right \} _{t=0}^{\infty}$ be a supermartingale with respect to
	$\left \{ Y_{t} \right \}_{t=0}^{\infty}$ and $T$ be a stopping time relating to $\left \{ Y_{t} \right \} _{t=0}^{\infty}$,}
\begin{align*}
	E(S_{T}|F_{0})\le S_{0}. 
\end{align*}

For convenience, we assume that for $\forall r_{i}\in S$, $0<\alpha\le r_{i}-r_{i-1}\le \beta$. Based on expected multiple-gain, the worst-case upper bound of EFHT can be estimated using Theorem 1.

\textbf{\emph{Theorem 1: }}{Let $\left\{f(X_{t})\right\}_{t=0}^{\infty}$ denote a stochastic process, where $f(X_{t})\ge 0$ holds for all $t\ge 0$. Assume that $f(X_{t})=r_{i}$, where $0<i\le m$, $t<T_{0}$. If there exists $k\in \mathbb{N}^+$ such that $G(t,k)=E(f(X_{t})-f(X_{t+k})|H_{t})\ge r_{i}-r_{i-1}\ge \alpha$ holds for all $t<T_{0}$, }
\begin{align*}
	E(T_{0}|f(X_{0}))\le \frac{kf(X_{0})}{\alpha}. 
\end{align*}

\emph{Proof:} Define $Q_{t}=f(X_{kt})+\frac{t}{k}\alpha, t=0, 1, 2, \dots$. Let $H_{kt}=\sigma (f(X_{0}), f(X_{k}), \dots, f(X_{kt}))$, for $\forall kt<T_{0}$,
\begin{align*}
	E(Q_{t+1}-Q_{t}|H_{kt})&=E(f(X_{kt+k})-f(X_{kt})+\frac{\alpha}{k} |H_{kt})\\
	&=E(f(X_{kt+k})-f(X_{kt})|H_{kt})+\frac{\alpha}{k}\\
	&=-G(kt,k)+\frac{\alpha}{k}\\
	&\le -\alpha+\frac{\alpha}{k}\\
	&\le 0. 
\end{align*}

This means that
\begin{align*}
	E(Q_{t+1}|H_{kt})&\le E(Q_{t}|H_{kt})\\
	&=Q_{t}. 
\end{align*}

\begin{spacing}{1}
	By Definition 1, $\left\{Q_{t}\right\}_{t=0}^{\infty}$ is a supermartingale relating to $\left\{f(X_{kt})\right\}_{t=0}^{\infty}$. Consider $T_{0}$ is a stopping time relating to stochastic process$\left \{ f(X_{kt}) \right \} _{t=0}^{\infty}$. By Lemma 1, we have 
	\begin{center}
		$E(Q_{T_{0}}|H_{0})\le Q_{0}=f(X_{0})$. 
	\end{center}
\end{spacing}

Therefore,
\begin{align*}
	f(X_{0})
	&\ge E(Q_{T_{0}}|H_{0})\\
	&=E(f(X_{kT_{0}})+\frac{T_{0}}{k}\alpha|H_{0})\\
	&=E(f(X_{kT_{0}})|H_{0})+\frac{\alpha E(T_{0}|H_{0})}{k}\\
	&\ge \frac{\alpha E(T_{0}|H_{0})}{k}.
\end{align*}

\begin{spacing}{1}
	This means that $\displaystyle E(T_{0}|H_{0})=E(T_{0}|f(X_{0}))\le \frac{kf(X_{0})}{\alpha}. $ $\hfill{\blacksquare}$
\end{spacing}

\begin{spacing}{1}
	The conclusion of Theorem 1 in this paper differs from that of Theorem 1 in \cite{yushan2016first}, as the former is based on expected multiple-gain, while the latter relies on average gain. Moreover, while the lower bound of average gain needs to be a constant in \cite{yushan2016first}, the lower bound of the expected multiple-gain depends on the target space $S=\left\{r_{0}, r_{1}, \dots, r_{m}\right\}$ in this paper.

	Theorem 1 describes a scenario where the fitness value changes only after every $k$ iterations, and the magnitude of the change is the smallest possible increment $\alpha$. Notably, the value of $k$ is related to the longest duration during which the gain remains zero in the execution process of EAs. Therefore, the conclusion of Theorem 1 represents the worst-case upper bound of EFHT. Furthermore, a high value of $k$ can lead to a less accurate worst-case upper bound on $E(T_{0})$. To improve the accuracy of the worst-case upper bound on $E(T_{0})$, it is necessary to obtain the lower bound of $k$. The definition of $k_{low}$ is presented below.
\end{spacing}

\textbf{\emph{Definition 6: }}{Assume that $f(X_{t})=r_{i}$, where $0<i\le m$, $t<T_{0}$. $k_{low}\in\mathbb{N}^+$ is the lower bound of $k$ if $G(t,k)\ge r_{i}-r_{i-1}$ holds for all $t<T_{0}$, $k\ge k_{low}$.}

We will discuss how to obtain $k_{low}$ in the next section. If we do not restrict the value of $k$, the average-case upper bound of EFHT can be estimated using Theorem 2.

\textbf{\emph{Theorem 2: }}{Let $\left \{ f(X_{t}) \right \}_{t=0}^{\infty}$ denote a stochastic process, where $f(X_{t})\ge 0$ holds for all $t\ge 0$. Assume that $f(X_{0})=r_{L}$, where $0<L\le m$. If there exists a monotonically non-decreasing function $h(r):\left\{0,1\right\}^{n}\rightarrow R^{+}$ and $k\in \mathbb{N}^+$ such that $G(t,k)=E(f(X_{t})-f(X_{t+k})|H_{t})\ge h(f(X_{t}))$ holds for all $t<T_{0}$, }
\begin{align*}
	E(T_{0}|f(X_{0}))\le k\cdot \sum_{i=1}^{L}\frac{r_{i}-r_{i-1}}{h(r_{i})}. 
\end{align*}

\emph{Proof:} Let $f(X_{t})=r_{p}$, $f(X_{t+k})=r_{q}$, where $0\le q<p<L$. 
Let $g(f(X_{t}))=\begin{cases}
	0,f(X_{t})=0\\
	\displaystyle \sum_{i=1}^{p}\frac{r_{i}-r_{i-1}}{h(r_{i})},f(X_{t})>0
\end{cases}$, for $\forall t<T_{0}$,

(1) When $t<T_{0}$, $t+k<T_{0}$,
\begin{align*}
	g(f(X_{t}))-g(f(X_{t+k}))&=\sum_{i=1}^{p}\frac{r_{i}-r_{i-1}}{h(r_{i})}-\sum_{i=1}^{q}\frac{r_{i}-r_{i-1}}{h(r_{i})}\\
	&=\sum_{i=q+1}^{p}\frac{r_{i}-r_{i-1}}{h(r_{i})}.
\end{align*}

Therefore,
\begin{align*}
	E\left[g(f(X_{t}))-g(f(X_{t+k}))|H_{t}\right]
	&=E(\sum_{i=q+1}^{p}\frac{r_{i}-r_{i-1}}{h(r_{i})}|H_{t})\\
	&\ge E(\frac{r_{p}-r_{q}}{h(r_{p})}|H_{t})\\
	&=E(\frac{f(X_{t})-f(X_{t+k})}{h(f(X_{t}))}|H_{t})\\
	&=\frac{G(t,k)}{h(f(X_{t}))}. 
\end{align*}

According to the assumption in Theorem 2, there exists a monotonically non-decreasing function $h(r)$ that satisfies $G(t,k)\ge h(f(X_{t}))$. Therefore,
\begin{align*}
	E\left[g(f(X_{t}))-g(f(X_{t+k}))|H_{t}\right]\ge 1. 
\end{align*}

(2) When $t<T_{0}$, $t+k\ge T_{0}$,
\begin{align*}
	g(f(X_{t}))-g(f(X_{t+k}))&=\sum_{i=1}^{p}\frac{r_{i}-r_{i-1}}{h(r_{i})}.
\end{align*}

Therefore,
\begin{align*}
	E\left[g(f(X_{t}))-g(f(X_{t+k}))|H_{t}\right]
	&=E(\sum_{i=1}^{p}\frac{r_{i}-r_{i-1}}{h(r_{i})}|H_{t})\\
	&\ge E(\sum_{i=q+1}^{p}\frac{r_{i}-r_{i-1}}{h(r_{i})}|H_{t})\\
	&\ge 1
\end{align*}

Note that $T_{0}=min\left\{t\ge 0:f(X_{t})=0\right\}=min\left\{t\ge 0:g(f(X_{t}))=0\right\}\stackrel{\wedge}{=}T_{0}^{g}$. According to Theorem 1, we have
\begin{align*}
	E(T_{0}|f(X_{0}))=E\left[T_{0}^{g}|g(f(X_{0}))\right]
	&\le \frac{k\cdot g(f(X_{0}))}{1}.
\end{align*}

\begin{spacing}{1}
	This means $E(T_{0}|f(X_{0}))\le k\cdot \sum_{i=1}^{L}\frac{r_{i}-r_{i-1}}{h(r_{i})}$. $\hfill\blacksquare$
\end{spacing}

\begin{spacing}{1}
	The conclusion of Theorem 2 in this paper differs from that of Theorem 2 in \cite{yushan2016first}, as the former is established based on expected multiple-gain, while the latter is built upon average gain. Furthermore, the lower bound of average gain must be a monotonically increasing and integrable function in Theorem 2 of \cite{yushan2016first}, while the lower bound of expected multiple-gain is a monotonically non-decreasing function in Theorem 2 of this paper. In addition, Theorem 2 in this paper can be applied to combinatorial optimization, while Theorem 2 in \cite{yushan2016first} can be applied to continuous optimization.

	The formula in Theorem 2 is determined by the initial solution $X_{0}$, the lower bound $h(r_{p})$ of $G(t,k)$, the parameter $k$ of $G(t,k)$ and the target space $S$. In addition, $G(t,k)\ge h(r_{p})$ indicates that the fitness value may not necessarily decrease after $k$ iterations, because $h(r_{p})$ is not always greater than $r_{p}-r_{p-1}$. Therefore, $k$ can be any positive integer if there exists a function $h(r_{p})$ that satisfies $G(t,k)\ge h(r_{p})$. This means that the conclusion of Theorem 2 represents the average-case upper bound of EFHT. When $h(r_{p})\ge r_{p}-r_{p-1}$, a special case can be derived as follows.
\end{spacing}

\textbf{\emph{Corollary 1: }}{Let $\left \{ f(X_{t}) \right \}_{t=0}^{\infty}$ be a stochastic process, where $f(X_{t})\ge 0$ holds for all $t\ge 0$. Assume that $f(X_{0})=r_{L}$, $f(X_{t})=r_{p}$, where $0<p\le m$, $t<T_{0}$. If there exists $k\in \mathbb{N}^+$ such that $G(t,k)=E(f(X_{t})-f(X_{t+k})|H_{t})\ge \alpha$ holds for all $t \ge 0$, }
\begin{align*}
	E(T_{0}|f(X_{0}))\le k\cdot \frac{r_{L}-r_{0}}{\alpha}. 
\end{align*}

\emph{Proof:} Let $h(f(X_{t}))=\alpha$. According to Theorem 2 and the assumption in Corollary 1, we have
\begin{align*}
	E(T_{0}|f(X_{0})=r_{L})&\le k\cdot \sum_{i=1}^{L}\frac{r_{i}-r_{i-1}}{\alpha}\\
	&= k\cdot \frac{r_{1}-r_{0}+r_{2}-r_{1}+\dots+r_{L}-r_{L-1}}{\alpha}
\end{align*}

It means that $E(T_{0}|f(X_{0})=r_{L})\le k\cdot \frac{r_{L}-r_{0}}{\alpha}$. $\hfill{\blacksquare}$

Note that Corollary 1 and Theorem 1 are equivalent when $r_{0}=0$. Theorem 2 will be utilized in Section III to analyze the average-case upper bounds of EFHT. The following section will present a detailed running-time analysis of evolutionary combinatorial optimization instances using the multiple-gain model.

\section{Running-time Analysis of EAs Based on Multiple-gain Estimation}

\begin{figure*}
	\centering
	\includegraphics[width=\textwidth]{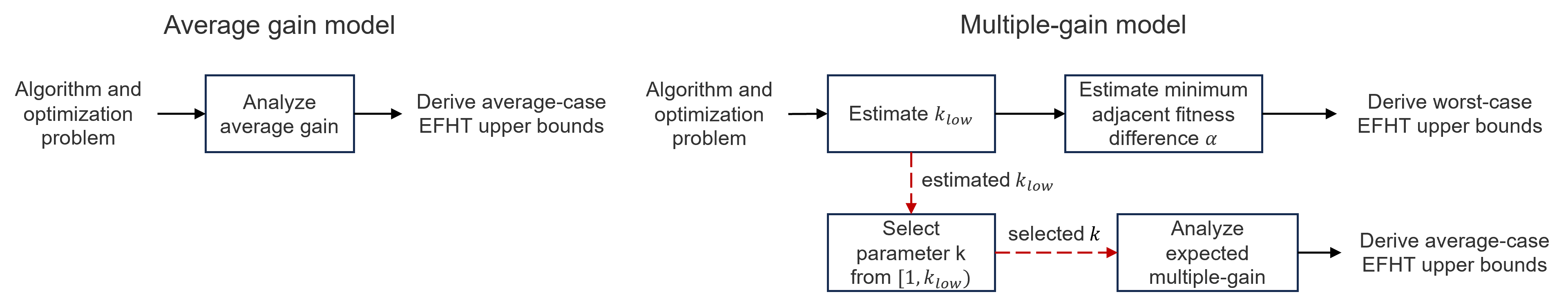}
	\caption{Flowchart of the average gain model and multiple-gain model}
\end{figure*}

Fig. 1 illustrates the overall procedures of both the average gain model and the multiple-gain model. The multiple-gain model estimates the worst-case upper bounds of EFHT through two additional steps, supported by experimental data. Furthermore, the multiple-gain model needs to select parameter $k$ before estimating the average-case upper bounds of EFHT compared to the average gain model.

\subsection{Multiple-gain Estimation for EFHT upper bounds of EAs}
The precise estimation of $k_{low}$ is crucial to the application of Theorem 1. According to Definition 6, it can be known that $k_{low}$ corresponds to the longest duration during which the gain remains zero. However, such duration obtained in each execution process may be different due to the randomness of EAs. These different durations will yield distinct worst-case upper bounds of EFHT. Therefore, we will utilize Corollary 2 to obtain a unified worst-case upper bound of EFHT.

\textbf{\emph{Corollary 2: }}{Let $\left\{f(X_{t})\right\}_{t=0}^{\infty}$ be a stochastic process, where $f(X_{t})\ge 0$ holds for all $t\ge 0$. Assume that $f(X_{t})=r_{i}$, where $0<i\le m$, $t<T_{0}$. If there exists $k\in \mathbb{N}^+$ such that $G(t,k)=E(f(X_{t})-f(X_{t+k})|H_{t})\ge r_{i}-r_{i-1}\ge \alpha$ holds for all $t<T_{0}$, }
\begin{align*}
	E(T_{0}|f(X_{0}))\le \frac{k_{low}f(X_{0})}{\alpha}.  
\end{align*}

\emph{Proof:} By Definition 6, we have
\begin{align*}
	E(f(X_{t})-f(X_{t+k_{low}})|H_{t})\ge r_{i}-r_{i-1}\ge \alpha
\end{align*}

According to Theorem 1, we have
\begin{align*}
	E(T_{0}|f(X_{0}))\le \frac{k_{low}f(X_{0})}{\alpha}. 
\end{align*}

$\hfill{\blacksquare}$

According to the law of large numbers, the larger the sample size, the higher the probability that the arithmetic mean will approach the expected value\cite{durrett2019}. Therefore, we can estimate $k_{low}$ by taking the average of each longest duration during which the gain remains zero obtained from repeated EAs.

\begin{spacing}{1}
	To utilize Corollary 2, we still need $\alpha$, which represents the minimum difference between adjacent fitness in the target space. If it is difficult to analyze the distribution of the target space, $\alpha$ can be estimated through experimental results. 
	Specifically, we can record all data of gain that are not equal to zero in the experiment. There will be at least one gain equal to $\alpha$ with a sufficiently large number of experiments. Therefore, we can select the minimum data of gain as the estimation of $\alpha$. Once the estimated $k_{low}$ and $\alpha$  are obtained, the worst-case upper bound of EFHT can be estimated using the formula $\frac{k_{low}f(X_{0})}{\alpha}$ presented in Corollary 2.
\end{spacing}

Before deriving the average-case upper bound of EFHT, it is necessary to select the parameter $k$ for expected multiple-gain, where the range of $k$ depends on the estimation of $k_{low}$ obtained from experiments. The
expected multiple-gain varies with different $k\in \left[1,k_{low}\right)$,
leading to different upper bounds of EFHT. However, recalculating the expected multiple-gain for each distinct $k$ is computationally intensive. Furthermore, the primary purpose of this paper is not to calculate different expected multiple-gains obtained from different values of $k$. Therefore, we set $k=1$ to compute the expected multiple-gain in the following subsection. For all the problems discussed in the next subsection, the state of individual $X_{t+1}$ depends only on the individual $X_{t}$. Therefore, $\left\{f(X_{t})\right\}_{t=0}^{\infty}$ can be modeled as a Markov chain. Consequently, the expected multiple-gain can be rewritten as
\begin{align*}
	G(t, 1)=E(f(X_{t})-f(X_{t+1})|f(X_{t})). 
\end{align*}


In the following three subsections, we utilize Theorem 2 to analyze the average-case upper bounds of EFHT for three evolutionary combinatorial optimization instances. Moreover, we conduct analyses to determine theoretical $k_{low}$ that satisfies Theorem 1 for these three evolutionary combinatorial optimization instances. These analyses validate the applicability of the multiple-gain model for different evolutionary combinatorial optimization instances.

\subsection{Running-time Analysis of  (1+1) EA for the Onemax Problem}

We first apply the multiple-gain model to analyze the EFHT of (1+1) EA for the Onemax problem. Although it is a simple instance, it still holds theoretical value. Additionally, this instance can be used to verify the correctness of the multiple-gain model. The Onemax problem is to maximize the number of one-bits in a binary string. Therefore, the global optimum is $X_{opt}=\left\{1,1,\dots,1\right\}$. Formally, for $X\in \left\{0, 1\right\}^{n}$,
\begin{align*}
	max\quad f(X)=\sum_{i=1}^{n}x_{i}\in S
\end{align*}
where $S=\left\{0,1,\dots,n\right\}$.

Detailed operations of (1+1) EA are described as follows. 
\begin{algorithm}
	\caption{(1+1) EA}\label{alg:(1+1) EA}
	\begin{algorithmic}[1]
		\REQUIRE {Encoding length $n$.}
		\ENSURE {The global optimum $X_{opt}$.}
		\STATE {Initialization: set generation $t$=0, generate a solution 
			$X_{0}= \left\{x_{0, 1}, x_{0, 2}, \dots x_{0, n}\right\}\in \left\{0, 1\right\}^{n}$ at random.}
		\WHILE{termination condition is not fulfilled}
		\STATE $y\leftarrow$ flip each bit of $X_{t}$ independently with probability $\frac{1}{n}$.
		\IF{$f\left(y\right) \ge f\left(X_{t}\right)$}
		\STATE $X_{t + 1} \leftarrow y$
		\ELSE 
		\STATE $X_{t + 1} \leftarrow X_{t}$
		\ENDIF
		\STATE $t \leftarrow t+1$
		\ENDWHILE
		\RETURN $X_{opt}$
	\end{algorithmic}
\end{algorithm}

Various conclusions have been derived regarding the running time of EAs for the Onemax problem\cite{giessen2017interplay,oliveto2011simplified, doerr2014impact,droste1998rigorous,adak2024runtime}. The derived bound of EFHT in Theorems 3 is consistent with the known asymptotic bounds $O(nlogn)$\cite{droste1998rigorous}.

\textbf{\emph{Theorem 3: }}{Let $Y_{t}=n-f(X_{t})$ and $T_{0}=min\left\{t\ge 0:f(X_{t})=0\right\}$, the EFHT of (1+1) EA for the Onemax problem satisfies}
\begin{align*}
	E(T_{0}|Y_{0}=r_{L})\le e\cdot n\sum_{x=1}^{n}\frac{1}{x}. 
\end{align*}

\emph{Proof:} Let $Y_{t}=r$ for $\forall t<T_{0}$,
\begin{align*}
	G(t, 1)&=E(Y_{t}-Y_{t+1}|Y_{t}=r)\\
	&=\sum_{i=1}^{r}i\cdot P(Y_{t}-Y_{t+1}=i|Y_{t}=r)\\
	&\ge P(Y_{t}-Y_{t+1}=1|Y_{t}=r)\\
	&\ge C_{r}^{1}\cdot \frac{1}{n}\cdot(1-\frac{1}{n})^{n-1}\\
	&\ge\frac{r}{e\cdot n}. 
\end{align*}

Let $h_{1}(r)=\frac{r}{en}$. Note that $h_{1}(r)$ is a monotonically non-decreasing function. According to Theorem 2, we have
\begin{align*}
	E(T_{0}|Y_{0}=r_{L})&\le e\cdot n\sum_{i=1}^{L}\frac{r_{i}-r_{i-1}}{r_{i}}\\
	&=e\cdot n(\frac{1-0}{1}+\frac{2-1}{2}+\dots+\frac{L-(L-1)}{L})\\
	&\le e\cdot n\sum_{x=1}^{n}\frac{1}{x}. 
\end{align*}

When $n\rightarrow \infty$,
\begin{align*}
	\lim_{n \to \infty}e\cdot n\sum_{x=1}^{n}\frac{1}{x}
	&=\lim_{n \to \infty}e\cdot nlogn 
\end{align*}

This means $E(T_{0}|Y_{0}=r_{L})\in O(nlogn)$. $\hfill{\blacksquare}$

\begin{spacing}{1}
	As demonstrated by the proof process of Theorem 3, the precise estimation of $G(t,1)$ is the vital part of Theorem 2. In this case, $r_{i}-r_{i-1}=1$. Moreover, $G(t, 1) \ge h_{1}(r)= \frac{r}{ne}$. Therefore, to achieve an expected multiple-gain of at least $r_{p}-r_{p-1}$ after $k$ iterations, $k$ must satisfy the condition $k \cdot h_{1}(r) \ge r_{p}-r_{p-1}$. 
\end{spacing}

\textbf{\emph{Theorem 4: }}{The $k_{low}$ of (1+1) EA for the Onemax problem is}
\begin{align*}
	k_{low}=en.
\end{align*}

\emph{Proof:} By the inequality $k \cdot h_{1}(r) \ge r_{p}-r_{p-1}=1$, we have
\begin{align*}
	k\ge \frac{1}{h_{1}(r)}=\frac{en}{r}
\end{align*}

Therefore, $k_{low}=en$.$\hfill\blacksquare$

Theorem 3 provides a series form average-case upper bound of EFHT that is consistent with the known asymptotic upper bound in \cite{droste1998rigorous}, and its proof is more concise. In the following subsection, we will conduct analyses of the knapsack problem with favorably correlated weights and general $k$-MAX-SAT problem by following a proof process analogous to that of Theorems 3 and 4.

\subsection{Running-time Analysis of (1+$\lambda$) EA for the Knapsack Problem with Favorably Correlated Weights}

The knapsack problem is one of the most important optimization problems. Frank et al. \cite{neumann2018running-time, neumann2019running-time, xie2021running-time} analyzed the running time of solving specific instances of the knapsack problem using (1+1) EA, GSEMO, and RLS. The knapsack problem is to maximize a linear profit function while satisfying a linear weight constraint \cite{neumann2019running-time}. For $X\in \left\{0, 1\right\}^{n}$, the knapsack problem is typically formulated as follows:
\begin{align*}
	max\quad f(X)=\sum_{i=1}^{n}v_{i}x_{i}\\
	s. t \quad B(X)=\sum_{i=1}^{n}w_{i}x_{i}\le K
\end{align*}
where $w_{i}>0$ denotes the weight of item $i$, $v_{i}>0$ denotes the value of item $i$, and $K$ represents the capacity of knapsack.

If $x_i$ is set to 1, this indicates that the item $i$ is to be included in the knapsack. Conversely, if $x_i$ is set to 0, the item $i$ will be excluded from the knapsack. Detailed operations of (1+$\lambda$) EA are described as follows. 

\begin{algorithm}
	\caption{(1+$\lambda$) EA}\label{alg:(1+lambda)EA}
	\begin{algorithmic}[1]
		\REQUIRE {Encoding length $n$.}
		\ENSURE {The global optimum $X_{opt}$.}
		\STATE {Initialization: set generation $t$=0, generate a solution 
			$X_{0}= \left\{x_{0, 1}, x_{0, 2}, \dots, x_{0, n}\right\}\in \left\{0, 1\right\}^{n}$ at random.}
		\WHILE{termination condition is not fulfilled}
		\STATE $y_{max} \leftarrow X_{t}$
		\FOR{$i=1$ to $\lambda$}
		\STATE $y_{t,i}\leftarrow$ flip each bit of $X_{t}$ independently with probability $\frac{1}{n}$.
		\IF{$B(y_{t,i})>k$}
		\STATE $y_{t,i} \leftarrow X_{t}$
		\ELSIF{$f(y_{t,i})>f(y_{max})$}
		\STATE $y_{max}\leftarrow y_{t,i}$
		\ENDIF
		\ENDFOR
		\STATE $X_{t+1} \leftarrow y_{max}$
		\STATE $t \leftarrow t+1$
		\ENDWHILE
		\RETURN $X_{opt}$
	\end{algorithmic}
\end{algorithm}

\begin{spacing}{1}
	We consider an instance of the knapsack problem where for two different items $i$ and $j$, the conditions $v_{i}\ge v_{j}$ and $w_{i}\le w_{j}$ hold for all $i<j$. This instance represents a generalization of the linear function optimization problem with a uniform constraint, as previously explored in\cite{friedrich2017analysis}. Neumann et al.\cite{neumann2018running-time} proved that the time complexity upper bound of (1 + 1) EA for this instance is $O(n^{2}(logn+p_{max}))$, where $p_{max}$ is the largest profit of given items. Theorem 5 gives a more general upper bound of EFHT of (1+$\lambda)$EA for this instance.

	Let $d_{min}=min_{v_{i}\ne v_{j}}\left\{|v_{i}-v_{j}|\right\}$, $v_{min}=min\left\{v_{i}|1\le i\le n\right\}$. $d_{min}$ and $v_{min}$ will be utilized in the subsequent theorems.
\end{spacing}

\textbf{\emph{Theorem 5: }}{Suppose $X_{opt}=\left \{\underset{q}{\underbrace{1,1,\dots,1}},0,0,\dots,0 \right \}$. Let $Y_{t}=r_{m}-f(X_{t})$ and $T_{0}=min\left\{t\ge 0:Y_{t}=0\right\}$, the EFHT of (1+$\lambda$) EA for the knapsack problem with favorably
	correlated weights satisfies}
\begin{align*}
	E(T_{0}|Y_{0}=r_{L})&\le \frac{r_{L}-r_{0}}{p(\varepsilon_{1}) d_{min} p_{low1} + p(\varepsilon_{2}) v_{min} p_{low2}}
\end{align*}
where $p(\varepsilon_{1})=1-p(\varepsilon_{2})$, $p(\varepsilon_{2})=(\sum_{i=0}^{q-1}C_{q}^{i})/N$, $p_{low1}=(1-e^{-\frac{\lambda}{n^{2}e}})$, $p_{low2}=(1-e^{-\frac{\lambda}{ne}})$, and $N$ denotes the number of feasible solutions.

\emph{Proof:} Let $N$ denote the number of feasible solutions. There is at least one 0-bit in the first $q$ bits for any $X_{t}$ where $t<T_{0}$. Therefore, the proof will be discussed in two cases. One case considers the scenario with at least one 1-bit in the last $n-q$ bits of $X_{t}$. The other case addresses the scenario where the last $n-q$ bits of $X_{t}$ are all 0-bits.

\textbf{Case 1}: There is at least one item $i$ such that $x_{t, i}=1$, $t<T_{0}$ and $q<i\le n$. Since $t<T_{0}$, there must exist an item $j$ such that $1\le j \le q, x_{t, j}=0$. Therefore, $X_{t}$ can be optimized by flipping $x_{t,i}$ and $x_{t,j}$. The probability that flipping $x_{t,i}$ and $x_{t,j}$ at least once among $\lambda$ offspring is 
\begin{align*}
	p_{1}=1-\left[1-(\frac{1}{n})^{2}(1-\frac{1}{n})^{n-2}\right]^{\lambda}. 
\end{align*}

Therefore,
\begin{align*}
	E(Y_{t}-Y_{t+1}|Y_{t}=r_{L})&\ge (v_{i}-v_{j})\cdot p_{1}\\
	&\ge d_{min}\cdot p_{1}. 
\end{align*}

\textbf{Case 2}: There does not exist an item $i$ such that $x_{t, i}=1$, $t<T_{0}$, and $q<i\le n$. Since $t<T_{0}$, there must exist an item $j$ such that $1\le j \le q$ and $x_{t, j}=0$. Therefore, $X_{t}$ can be optimized by flipping $x_{t,j}$. The probability that flipping $x_{t,j}$ at least once among $\lambda$ offspring is 
\begin{align*}
	p_{2}=1-\left[1-(\frac{1}{n})^{1}(1-\frac{1}{n})^{n-1}\right]^{\lambda}. 
\end{align*}

Therefore
\begin{align*}
	E(Y_{t}-Y_{t+1}|Y_{t}=r_{L})&\ge v_{i}\cdot p_{2}\\
	&\ge v_{min}\cdot p_{2}. 
\end{align*}

Let $\varepsilon_{1}$ denote Case 1 and $\varepsilon_{2}$ denote Case 2. The probability of $\varepsilon_{1}$ occurring is $p(\varepsilon_{1})=1-p(\varepsilon_{2})$. Note that Case 2 represents the situation where the last $n-q$ bits of $X_{t}$ are all 0-bits. Furthermore, there are at most $q-1$ bits in the first $q$ bits of $X_{t}$ are 1-bits. Therefore, the probability of $\varepsilon_{2}$ occurring is $p(\varepsilon_{2})=(\sum_{i=0}^{q-1}C_{q}^{i})/N$. Consequently, we can obtain the following inequality:
\begin{align*}
	G(t,1)&=E(Y_{t}-Y_{t+1}|Y_{t})\\
	&=p(\varepsilon_{1})E(Y_{t}-Y_{t+1}|\varepsilon_{1})+p(\varepsilon_{2})E(Y_{t}-Y_{t+1}|\varepsilon_{2})\\
	&\ge p(\varepsilon_{1})d_{min}p_{1} + p(\varepsilon_{2})v_{min}p_{2}. 
\end{align*}

By the inequality
\begin{align*}
	(1-\frac{1}{n})^{n-2}>(1-\frac{1}{n})^{n-1}\ge \frac{1}{e},
\end{align*}

we have
\begin{align*}
	1-(\frac{1}{n})^{2}(1-\frac{1}{n})^{n-2}\le 1-\frac{1}{n^{2}e}
\end{align*}

and
\begin{align*}
	1-(\frac{1}{n})^{1}(1-\frac{1}{n})^{n-1}\le 1-\frac{1}{ne}. 
\end{align*}

When $0<a\le b$ and $n>0$, we have $a^{n}\le b^{n}$. Furthermore, $(1+x)\le e^{ x}, \forall x\in R$. Therefore, when $x>-1$ and $\lambda>0$, we have,
\begin{align*}
	(1+x)^{\lambda}\le e^{\lambda x}
\end{align*}

As $-\frac{1}{n^{2}e}>-1$, $-\frac{1}{ne}>-1$, we have, 
\begin{align*}
	(1-\frac{1}{n^{2}e})^{\lambda}\le e^{-\frac{\lambda}{n^{2}e}}
\end{align*}

and
\begin{align*}
	(1-\frac{1}{ne})^{\lambda}\le e^{-\frac{\lambda}{ne}}. 
\end{align*}

Let $p_{low1}=1-e^{-\frac{\lambda}{n^{2}e}}$, $p_{low2}=1-e^{-\frac{\lambda}{ne}}$. Therefore,
\begin{align*}
	G(t,1)&\ge p(\varepsilon_{1}) d_{min} p_{1} + p(\varepsilon_{2}) v_{min} p_{2}\\
	&\ge p(\varepsilon_{1})d_{min}p_{low1} + p(\varepsilon_{2})v_{min}p_{low2}
\end{align*}

Let $h_{2}(r)=p(\varepsilon_{1}) d_{min} p_{low1} + p(\varepsilon_{2}) v_{min} p_{low2}$. Note that $h_{2}(r)$ is a constant that can be considered a monotonically non-decreasing function.  By Theorem 2, we have 
\begin{align*}
	E(T_{0}|Y_{0}=r_{L})&\le \sum_{i=1}^{L}\frac{r_{i}-r_{i-1}}{p(\varepsilon_{1}) d_{min} p_{low1} + p(\varepsilon_{2}) v_{min} p_{low2}}\\
	&= \frac{r_{L}-r_{0}}{p(\varepsilon_{1}) d_{min} p_{low1} + p(\varepsilon_{2}) v_{min} p_{low2}}
\end{align*}

This means $E(T_{0}|Y_{0})\le \frac{r_{L}-r_{0}}{p(\varepsilon_{1}) d_{min} p_{low1} + p(\varepsilon_{2}) v_{min} p_{low2}}$. $\hfill{\blacksquare}$

Moreover, the conclusion of Theorem 5 can be discussed in two distinct cases, leading to the following corollary.

\textbf{\emph{Corollary 3: }}{Let $Y_{t}=r_{m}-f(X_{t})$ and $T_{0}=min\left\{t\ge 0:Y_{t}=0\right\}$, the EFHT of (1+$\lambda$) EA for the knapsack problem with favorably
	correlated weights satisfies}
\begin{align*}
	E(T_{0}|Y_{0}=r_{L})\in 
	\begin{cases}
		O(\frac{n(r_{L}-r_{0})}{\lambda v_{min}}),d_{min}p_{low1}-v_{min}p_{low2}\ge 0
		\\
		O(\frac{n^{2}(r_{L}-r_{0})}{\lambda d_{min}}),d_{min}p_{low1}-v_{min}p_{low2}< 0
	\end{cases}
\end{align*}
\emph{Proof:} By Theorem 5, we have
\begin{align*}
	E(T_{0}|Y_{0}=r_{L})
	&\le \frac{r_{L}-r_{0}}{p(\varepsilon_{1}) d_{min} p_{low1} + p(\varepsilon_{2}) v_{min} p_{low2}}\\
	&=\frac{r_{L}-r_{0}}{p(\varepsilon_{1})\left[d_{min}p_{low1}-v_{min}p_{low2}\right]+ v_{min}p_{low2}}
\end{align*}

(1) When $d_{min}p_{low1}-v_{min}p_{low2}\ge 0$,
\begin{align*}
	E(T_{0}|Y_{0}=r_{L})&\le \frac{r_{L}-r_{0}}{v_{min}p_{low2}}\\
	&=\frac{r_{L}-r_{0}}{v_{min}}\frac{1}{(1-e^{-\frac{\lambda}{ne}})}
\end{align*}

When $n\rightarrow \infty$,
\begin{align*}
	\lim_{n \to \infty} \frac{r_{L}-r_{0}}{v_{min}}\frac{1}{(1-e^{-\frac{\lambda}{ne}})}&=\lim_{n \to \infty}\frac{r_{L}-r_{0}}{v_{min}}\frac{1}{1-(1-\frac{\lambda}{ne})}\\
	&=\lim_{n \to \infty}\frac{ne(r_{L}-r_{0})}{\lambda v_{min}} 
\end{align*}

This means $E(T_{0}|Y_{0}=r_{L})\in O(\frac{n(r_{L}-r_{0})}{\lambda v_{min}})$.

(2) When $d_{min}p_{low1}-v_{min}p_{low2}<0$,
\begin{align*}
	E(T_{0}|Y_{0}=r_{L})
	&\le \frac{r_{L}-r_{0}}{d_{min}p_{low1}}\\
	&=\frac{r_{L}-r_{0}}{d_{min}}\frac{1}{(1-e^{-\frac{\lambda}{n^{2}e}})}
\end{align*}

When $n\rightarrow \infty$,
\begin{align*}
	\lim_{n \to \infty} \frac{r_{L}-r_{0}}{d_{min}}\frac{1}{(1-e^{-\frac{\lambda}{n^{2}e}})}&=\lim_{n \to \infty}\frac{r_{L}-r_{0}}{d_{min}}\frac{1}{1-(1-\frac{\lambda}{n^{2}e})}\\
	&=\lim_{n \to \infty}\frac{n^{2}e(r_{L}-r_{0})}{\lambda d_{min}} 
\end{align*}

\begin{spacing}{1}
	This means $E(T_{0}|Y_{0}=r_{L})\in O(\frac{n^{2}(r_{L}-r_{0})}{\lambda d_{min}})$. $\hfill{\blacksquare}$
\end{spacing}

\begin{spacing}{1}
	Note that $p(\varepsilon_{1})d_{min}p_{low1} + p(\varepsilon_{2})v_{min}p_{low2}$ equals a constant for a fixed $n$ and $\lambda$. This implies that the iterations needed for the gain to change do not depend on how far the current individual is from the global optimum. Moreover, $G(t, 1)\ge h_{2}(r_{p})$. Therefore, to achieve an expected multiple-gain of at least $r_{p}-r_{p-1}$ after $k$ iterations, $k$ must satisfy the condition $k\cdot h_{2}(r_{p})\ge r_{p}-r_{p-1}$.
\end{spacing}

\textbf{\emph{Theorem 6: }}{The $k_{low}$ of (1+$\lambda$) EA for the knapsack problem with favorably correlated weights is}
\begin{align*}
	k_{low}=\frac{\beta}{p(\varepsilon_{1}) d_{min} p_{low1} + p(\varepsilon_{2}) v_{min} p_{low2}}
\end{align*}
\emph{Proof:} Note that $\forall r_{i}\in S$, $0<\alpha\le r_{i}-r_{i-1}\le \beta$. By the inequality $k\cdot h_{2}(r_{p})\ge r_{p}-r_{p-1}$, we have 
\begin{align*}
	k\ge \frac{\beta}{h_{2}(r_{p})}
\end{align*}

Therefore, $k_{low}=\frac{\beta}{\left[p(\varepsilon_{1})d_{min}p_{low1} + p(\varepsilon_{2})v_{min}p_{low2}\right]}$. $\hfill{\blacksquare}$

\begin{spacing}{1}
	Theorem 5 presents a more general result than that found in \cite{neumann2018running-time}. Additionally, when $\lambda=1$ and $d_{min}p_{low1}-v_{min}p_{low2}<0$, Corollary 3 shows a new upper bound of EFHT $O(\frac{n^{2}(r_{L}-r_{0})}{d_{min}})$, which is tighter than the result $O(n^{2}(logn+p_{max}))$ in \cite{neumann2018running-time} by a factor of $logn$. Furthermore, a better upper bound of EFHT $O(\frac{n(r_{L}-r_{0})}{v_{min}})$ which is tighter than the result in \cite{neumann2018running-time} by a factor of $nlogn$ is also derived when $\lambda=1$ and $d_{min}p_{low1}-v_{min}p_{low2}\ge 0$.
\end{spacing}

\subsection{Running-time Analysis of (1+$\lambda$) EA for the $k$-MAX-SAT problem}

Existing running-time analysis studies of the SAT problem have focused on specific instances. For example, Buzdalov and Doerr et al. \cite{buzdalov2017running-time,doerr2015improved} analyzed the running time of EAs for solving random 3-CNF formulas. Sutton et al. \cite{sutton2012parameterized} conducted a running-time analysis of the 2-MAX-SAT problem. We will perform a running-time analysis of the general $k$-MAX-SAT problem to give a more comprehensive view of EAs on this problem.

A $k$-MAX-SAT problem instance is formulated as follows:
\begin{align*}
	l=\left\{(l_{1,1}\vee l_{1,2}\vee \dots l_{1,k}),\dots,(l_{s,1}\vee l_{s,2}\vee \dots l_{s,k})\right\}
\end{align*}
where the length of each clause is $k$ and $l_{i,j}$ denote a Boolean variable or its negation. The candidate solutions for the $k$-MAX-SAT instance are represented as binary strings of length $n$. Every bit of the binary string can be interpreted as the state of a Boolean variable $v_{i}$ (i.e., $x_{i}=1$ corresponds to $v_{i}=true$; $x_{i}=0$ corresponds to $v_{i}=false$). Let the function $f(X):\left\{0,1\right\}^{n}\rightarrow \left\{0,1,\dots,s\right\}$ count the clauses in $l$ that are satisfied under the assignment corresponding to $X$. Therefore, the
$k$-MAX-SAT problem is transformed into a pseudo-Boolean function optimization problem.

Let $N_{opt}$ denote the number of global optimums. Theorem 7 gives the upper bound of EFHT of (1+$\lambda$) EA with mutation probability $\frac{1}{2}$ for the $k$-MAX-SAT problem. In the (1+$\lambda$) EA, the mutation probability is conventionally set to $\frac{1}{n}$, as described in Algorithm 2. The reason for setting the mutation probability to $\frac{1}{2}$ is to streamline the analysis of expected multiple-gain. With a mutation probability of $\frac{1}{2}$, the probability of the current individual mutating into the global optimum is $\frac{N_{opt}}{2^{n}}$.

\textbf{\emph{Theorem 7: }}{Let $Y_{t}=f(X_{opt})-f(X_{t})$ and $T_{0}=min\left\{t\ge 0:Y_{t}=0\right\}$, the EFHT of (1+$\lambda$) EA with mutation probability $\frac{1}{2}$  for $k$-MAX-SAT problem satisfies}
\begin{align*}
	E(T_{0}|Y_{0})\le \frac{\sum_{x=1}^{s}\frac{1}{x}}{(1-e^{\frac{-\lambda\cdot N_{opt}}{2^{n}}})\cdot N_{opt}}. 	
\end{align*}

\emph{Proof:}\quad Assume that $Y_{t}=r$ where $t<T_{0}$. There are at least $N_{opt}$ solutions in the solution space that are superior to $X_{t}$. It follows that the probability of an offspring individual being superior to $X_{t}$ is at least $\frac{N_{opt}}{2^{n}}$. The probability that all of the $\lambda$ offspring individuals are inferior to $X_{t}$ is at most $(1-\frac{N_{opt}}{2^{n}})^{\lambda}$. Consequently, the probability that there is at least one individual among the $\lambda$ offspring individuals is superior to $X_{t}$ is greater than
\begin{align*}
	p_{3}=1-(1-\frac{N_{opt}}{2^{n}})^{\lambda}. 
\end{align*}

As $-\frac{N_{opt}}{2^{n}}>-1$, we have $p_{3}\ge 1-e^{\frac{-\lambda\cdot N_{opt}}{2^{n}}}$.

Therefore,
\begin{align*}
	G(t,1)&=E(Y_{t}-Y_{t+1}|Y_{t}=r)\\
	&\ge p_{3}\cdot N_{opt}\cdot r\\
	&\ge (1-e^{\frac{-\lambda\cdot N_{opt}}{2^{n}}})\cdot N_{opt}\cdot r. 
\end{align*}

Assume that $Y_{0}=L=r_{p}$, $f(X_{opt})=r_{q}\le s$, $r_{0}\ge 0$. Let $h_{3}(r)=(1-e^{\frac{-\lambda\cdot N_{opt}}{2^{n}}})\cdot N_{opt}\cdot r$. As $h_{3}(r)$ is a monotonically non-decreasing function, by Theorem 2, we have
\begin{align*}
	E(T_{0}|Y_{0}=r_{p})&\le \frac{\sum_{i=1}^{p}\frac{r_{i}-r_{i-1}}{r_{i}}}{(1-e^{\frac{-\lambda\cdot N_{opt}}{2^{n}}})\cdot N_{opt}}\\
	&\le \frac{\sum_{i=1}^{q}\frac{r_{i}-r_{i-1}}{r_{i}}}{(1-e^{\frac{-\lambda\cdot N_{opt}}{2^{n}}})\cdot N_{opt}}
\end{align*}

By the inequality
\begin{align*}
	\frac{r_{i}-r_{i-1}}{r_{i}}\le \frac{1}{r_{i-1}+1}+\frac{1}{r_{i-1}+2}+\dots+\frac{1}{r_{i}}
\end{align*}

we have
\begin{align*}
	\sum_{i=1}^{q}\frac{r_{i}-r_{i-1}}{r_{i}}\le \sum_{x=r_{0}+1}^{r_{q}}\frac{1}{x} \le \sum_{x=1}^{s}\frac{1}{x}
\end{align*}

This means $E(T_{0}|Y_{0})\le \frac{\sum_{x=1}^{s}\frac{1}{x}}{(1-e^{\frac{-\lambda\cdot N_{opt}}{2^{n}}})\cdot N_{opt}}$. $\hfill{\blacksquare}$

Moreover, the conclusion of Theorem 7 can lead to the following corollary.

\textbf{\emph{Corollary 4: }}{Let $Y_{t}=m-f(X_{t})$ and $T_{0}=min\left\{t\ge 0:Y_{t}=0\right\}$, the EFHT of (1+$\lambda$) EA with mutation probability $\frac{1}{2}$  for $k$-MAX-SAT problem satisfies}
\begin{align*}
	E(T_{0}|Y_{0}=r_{L})\in O(\frac{2^{n}}{\lambda}). 	
\end{align*}

\emph{Proof:}
According to Theorem 7, we have
\begin{align*}
	E(T_{0}|Y_{0}=r_{L})&\le \frac{\sum_{x=1}^{s}\frac{1}{x}}{(1-e^{\frac{-\lambda\cdot N_{opt}}{2^{n}}})\cdot N_{opt}}
\end{align*}
When $n\rightarrow \infty$,
\begin{align*}
	\lim_{n \to \infty} \frac{\sum_{x=1}^{s}\frac{1}{x}}{(1-e^{\frac{-\lambda\cdot N_{opt}}{2^{n}}})\cdot N_{opt}}&=\lim_{n \to \infty}\frac{\sum_{x=1}^{s}\frac{1}{x}}{\left[1-(1-\frac{\lambda\cdot N_{opt}}{2^{n}})\right]\cdot N_{opt}}\\
	&=\lim_{n \to \infty}\frac{\sum_{x=1}^{s}\frac{1}{x}}{\lambda}\frac{2^{n}}{(N_{opt})^{2}} 
\end{align*}
\begin{spacing}{1}
	It means that $E(T_{0}|Y_{0}=r_{L})\in O(\frac{2^{n}}{\lambda})$. $\hfill{\blacksquare}$
\end{spacing}

\begin{spacing}{1}
	Note that $G(t, 1)\ge h_{3}(r_{p})$. Therefore, to achieve an expected multiple-gain of at least $r_{p}-r_{p-1}$ after $k$ iterations, $k$ must satisfy the condition $k\cdot h_{3}(r_{p})\ge r_{p}-r_{p-1}$. 
\end{spacing}

\textbf{\emph{Theorem 8: }}{The $k_{low}$ of (1+$\lambda$) EA with mutation probability $\frac{1}{2}$ for the $k$-MAX-SAT problem is}
\begin{align*}
	k_{low}=\frac{\beta}{(1-e^{\frac{-\lambda\cdot N_{opt}}{2^{n}}})\cdot N_{opt}}.
\end{align*}
\emph{Proof:} Note that $\forall r_{i}\in S$, $0<\alpha\le r_{i}-r_{i-1}\le \beta$. By the inequality $k\cdot h_{3}(r_{p})\ge r_{p}-r_{p-1}$, we have 
\begin{align*}
	k\ge \frac{\beta}{h_{3}(r_{p})}
\end{align*}
\begin{spacing}{1}
	Therefore, $k_{low}=\beta/\left[(1-e^{\frac{-\lambda\cdot N_{opt}}{2^{n}}})\cdot N_{opt}\right]$. $\hfill{\blacksquare}$
\end{spacing}

Theorem 7 provides a first average-case upper bound of EFHT in closed-form expression of (1+$\lambda$)EA with mutation probability $\frac{1}{2}$ for any $k$-MAX-SAT instances. A thorough study of this specific class of (1+$\lambda$)EA with special mutation operator will help reveal the dynamic behavior of general (1+$\lambda$)EA.

According to Theorems 3, 5, and 7, the theoretical upper bounds of EFHT obtained
by the multiple-gain model are guaranteed to be greater than EFHT.
The correctness of the theoretical upper bounds of EFHT will be confirmed in Section IV.

\section{Experiments}
We will conduct the following three experiments to verify whether the experimental results align with the theorems in Section III. The experiments include (1+1) EA for the Onemax problem, (1+$\lambda$) EA for the knapsack problem with favorably correlated weights, and (1+$\lambda$) EA for the $k$-MAX-SAT problem.

\begin{spacing}{1}
	In the following three experiments, each algorithm is repeated 1000 times for each encoding length $n$. Let $T_{0i}$ denote the FHT of the $i$-th round, and $T_{max}$ denote the largest FHT in 1000 rounds. $\widehat{T_{0}}=\sum_{i=1}^{1000}T_{0i}/{1000}$ is considered to be the estimation of the actual
	EFHT. Let $k_{i}$ denote the most prolonged time interval during which the gain remains zero in the $i$-th round. $\widehat{k}=\sum_{i=1}^{1000}k_{i}/{1000}$ was considered to be the estimation of $k_{low}$. Let EFHT1 denote the theoretical average-case upper bound of EFHT. Let EFHT2 denote the theoretical worst-case upper bound of EFHT.
\end{spacing}

The correlation coefficient is commonly used to describe the relationship between variables. Therefore, we utilize the correlation coefficient to assess whether the experimental results align with the theorems in Section III. The formula for calculating the correlation coefficient $r(x,y)$ is as follow\cite{1997Statistics}:
\begin{align*}
	r(x,y)=\frac{n\sum xy-\sum x\sum y}{\sqrt{n\sum x^2-\left(\sum x\right)^2}\sqrt{n\sum y^2-\left(\sum y\right)^2}}
\end{align*}

As $|r(x,y)|$ approaches 1, it indicates a stronger correlation between $x$ and $y$, while $|r(x,y)|$ approaching $0$ signifies a weaker correlation. For each experiment, we will present the computed results for the three correlation coefficients: $r(EFHT1,T_{0})$, $r(EFHT2,T_{max})$, and $r(\widehat{k},k_{low})$. The criteria for evaluating the correlation coefficient may vary among researchers from different fields. However, a correlation coefficient greater than 0.91 generally indicates a strong relationship for almost all researchers\cite{1997Statistics}. Therefore, if EFHT1$>\widehat{T_{0}}$, EFHT2$>T_{max}$, $\widehat{k}>k_{low}$, $r(EFHT1,T_{0})>0.91$, $r(EFHT2,T_{max})>0.91$, and $r(\widehat{k},k_{low})>0.91$, we consider the experimental results to align with the theorems in Section III.

Commonly, calculating EFHT1, EFHT2 and the theoretical value of $k_{low}$ requires different data for each case. Since case analysis is not the primary focus of the experiments, we use self-constructed cases rather than the test dataset to simplify the calculation. Since we are using self-constructed cases, there is no need to estimate $\alpha$. In addition, we implement the algorithms as described in Section III in Matlab. 
\subsection{Verification of Theorems 3 and 4}
In this experiment, we will verify the correctness of Theorems 3 and 4. The experimental settings of Algorithm 1 for the Onemax problem are listed as follows: set the initial population $X_{0}=\left\{0, 0, \dots, 0\right\}$, which leads to $Y_{0}=n$; the target space $S=\left\{0,1,2,\dots,n\right\}$; for $\forall r_{p}\in S$, $1=\alpha\le r_{p}-r_{p-1}\le 1=\beta$; the encoding length is chosen from the set $\left\{10,11,12,\dots,30\right\}$. Fig. 2 shows the numerical results.

\begin{spacing}{1}
	Fig. 2(a) shows the EFHT1 and $\widehat{T_{0}}$ with respect to the encoding length where EFHT1 represents the theoretical upper bound $ne \cdot \sum_{x=1}^{L}\frac{1}{x}$ given by Theorem 3. The correlation coefficient between the EFHT1 and $T_{0}$ can be obtained as follows:
	\begin{align*}
		r_{1}(EFHT1,T_{0})=0.9999
	\end{align*}
\end{spacing}

Fig. 2(b) shows the EFHT2 and $T_{max}$ with respect to the encoding length where the EFHT2 represents the theoretical upper bound $\widehat{k}Y_{0}/\alpha$ given by Corollary 2. The correlation coefficient between EFHT2 and $T_{max}$ can be obtained as follows:
\begin{align*}
	r_{1}(EFHT2,T_{max})=0.9984
\end{align*}

\begin{spacing}{1}
	Fig. 2(c) shows the $\widehat{k}$ and $k_{low}$ with respect to the encoding length where $k_{low}$ equals $ne$ given by Theorem 4. The correlation coefficient between $\widehat{k}$ and $k_{low}$ can be obtained as follows:
	\begin{align*}
		r_{1}(\widehat{k},k_{low})=0.9993
	\end{align*}
\end{spacing}

From Fig. 2, it is evident that EFHT1$>\widehat{T_{0}}$, EFHT2$>T_{max}$, and $\widehat{k}>k_{low}$. Furthermore, $r_{1}(EFHT1,T_{0})$, $r_{1}(EFHT2,T_{max})$ and $r_{1}(\widehat{k},k_{low})$ are all greater than 0.91. Therefore, we conclude that the experimental results align with Theorem 3, Theorem 4, and Corollary 2.

\begin{figure*}
	\centering
	\subfloat[\label{a}]{
		\includegraphics[width=0.32\textwidth]{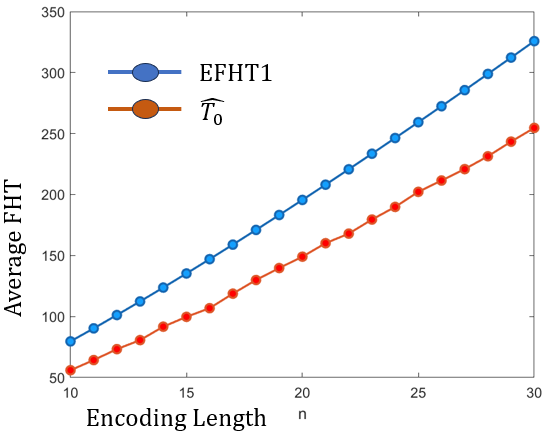}}
	\subfloat[\label{b}]{
		\includegraphics[width=0.32\textwidth]{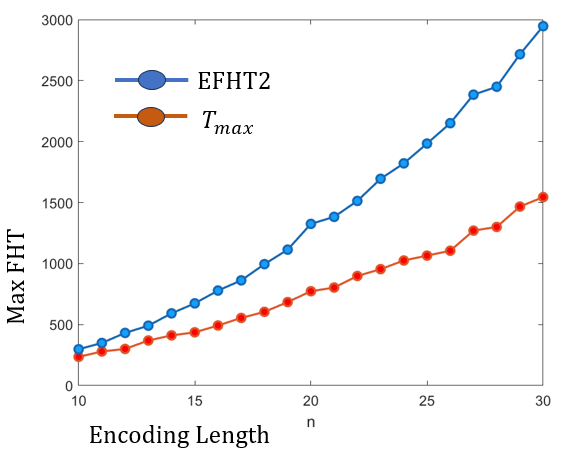}}
	\subfloat[\label{c}]{
		\includegraphics[width=0.32\textwidth]{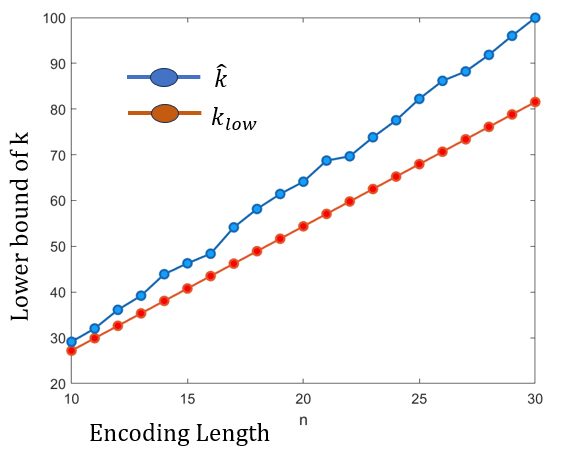}}
	\caption{Results of experiment A. (a) Theoretical average-case upper bounds of EFHT and estimation of actual EFHT. (b) Theoretical worst-case upper bounds of EFHT and actual largest FHT. (c) Estimation of $k_{low}$ and theoretical value of $k_{low}$ }
\end{figure*}

\subsection{Verification of Theorems 5 and 6}
In this experiment, we will verify the correctness of Theorems 5 and 6. The experimental settings of Algorithm 2 for the knapsack problem with favorably correlated weights are listed as follows: set the capacity of knapsack $K=3$, $\lambda=20$; the value of the first three items is set to 3, 3, 1, with weights of 1 each; the values of all the remaining items are set as 1, with weights of 2 each; the target space $S=\left\{0, 1, 2, \dots, 4,6,7\right\}$; for $\forall r_{p}\in S$, $1=\alpha\le r_{p}-r_{p-1}\le 2=\beta$; $d_{min}=2$, $v_{min}=1$, $q=3$, $N=C_{n}^{0}+C_{n}^{1}+C_{3}^{1}C_{n-3}^{1}+C_{3}^{2}+C_{3}^{3}$; set the initial population $X_{0}=\left\{0, 0, \dots, 0\right\}$, which leads to $Y_{0}=7$; the encoding length is chosen from the set $\left\{10,11,12,\dots,30\right\}$. Fig. 3 shows the numerical results.

\begin{figure*}
	\centering
	\subfloat[\label{a}]{
		\includegraphics[width=0.32\textwidth]{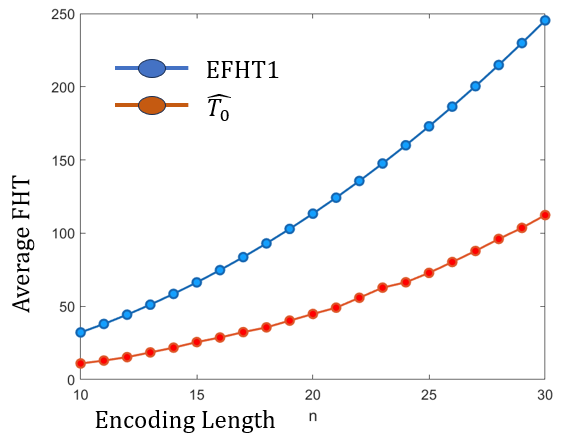}}
	\subfloat[\label{b}]{
		\includegraphics[width=0.32\textwidth]{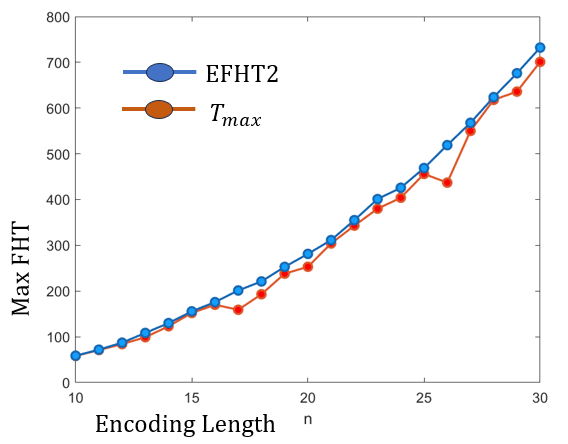}}
	\subfloat[\label{c}]{
		\includegraphics[width=0.32\textwidth]{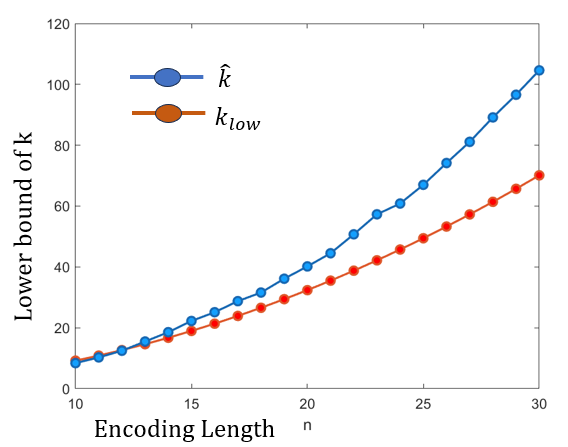}}
	\caption{Results of experiment B. (a) Theoretical average-case upper bounds of EFHT and estimation of actual EFHT. (b) Theoretical worst-case upper bounds of EFHT and actual largest FHT. (c) Estimation of $k_{low}$ and theoretical value of $k_{low}$}
\end{figure*}

\begin{spacing}{1}
	Fig. 3(a) shows the EFHT1 and $\widehat{T_{0}}$ with respect to the encoding length where EFHT1 represents the theoretical upper bound $(Y_{0}-0)/\left[p(\varepsilon_{1})\cdot d_{min}\cdot p_{low1} + p(\varepsilon_{2})\cdot v_{min}\cdot p_{low2}\right]$ given by Theorem 3. The correlation coefficient between EFHT1 and $T_{0}$ can be obtained as follows:
	\begin{align*}
		r_{2}(EFHT1,T_{0})=0.9985
	\end{align*}
\end{spacing}

Fig. 3(b) shows the EFHT2 and $T_{max}$ with respect to the encoding length where EFHT2 represents the theoretical upper bound $\widehat{k}Y_{0}/\alpha$ given by Corollary 2. The correlation coefficient between EFHT2 and $T_{max}$ can be obtained as follows:
\begin{align*}
	r_{2}(EFHT2,T_{max})=0.9965
\end{align*}

\begin{spacing}{1}
	Fig. 3(c) shows the $\widehat{k}$ and $k_{low}$ with respect to the encoding length where $k_{low}$ equals $\beta/\left[p(\varepsilon_{1})d_{min}p_{low1} + p(\varepsilon_{2})v_{min}p_{low2}\right]$ given by Theorem 6. The correlation coefficient between $\widehat{k}$ and $k_{low}$ can be obtained as follows:
	\begin{align*}
		r_{2}(\widehat{k},k_{low})=0.9983
	\end{align*}
\end{spacing}

In Fig. 3, EFHT1$>\widehat{T_{0}}$, EFHT2$>T_{max}$ and $\widehat{k}>k_{low}$. Additionally, $r_{2}(EFHT1,T_{0})$, $r_{2}(EFHT2,T_{max})$ and $r_{2}(\widehat{k},k_{low})$ are all greater than 0.91. Therefore, the experimental results align with Theorem 5, Theorem 6, and Corollary 2.

\subsection{Verification of Theorems 7 and 8}
In this experiment, we will verify the correctness of Theorems 7 and 8. The experimental settings of Algorithm 2 with mutation probability $\frac{1}{2}$ for $k$-MAX-SAT problem are listed as follows: set the length of each clause $k=2$, $\lambda=20$; $f(X)=(x_{1}\vee \neg x_{2} )\wedge\dots(x_{1}\vee \neg x_{n})\wedge(\neg x_{1}\vee x_{2} )\wedge\dots(\neg x_{1}\vee x_{n} )$; the global optimums of $f(X)$ are $(0, 0, \dots, 0)$ and $(1, 1, \dots, 1)$, thus $N_{opt}=2$; the target space $S=\left\{n-1, n, \dots, 2(n-1)\right\}$; for $\forall r_{p}\in S$, $1=\alpha\le r_{p}-r_{p-1}\le 1=\beta$; set the initial population $X_{0}=\left\{0, 1, 1, \dots, 1\right\}$, which leads to $Y_{0}=n-1$; the encoding length is chosen from the set $\left\{5,6,\dots,14,15\right\}$. Fig. 4 shows the numerical results.

\begin{figure*}
	\centering
	\subfloat[\label{fig:a}]{
		\includegraphics[width=0.33\textwidth]{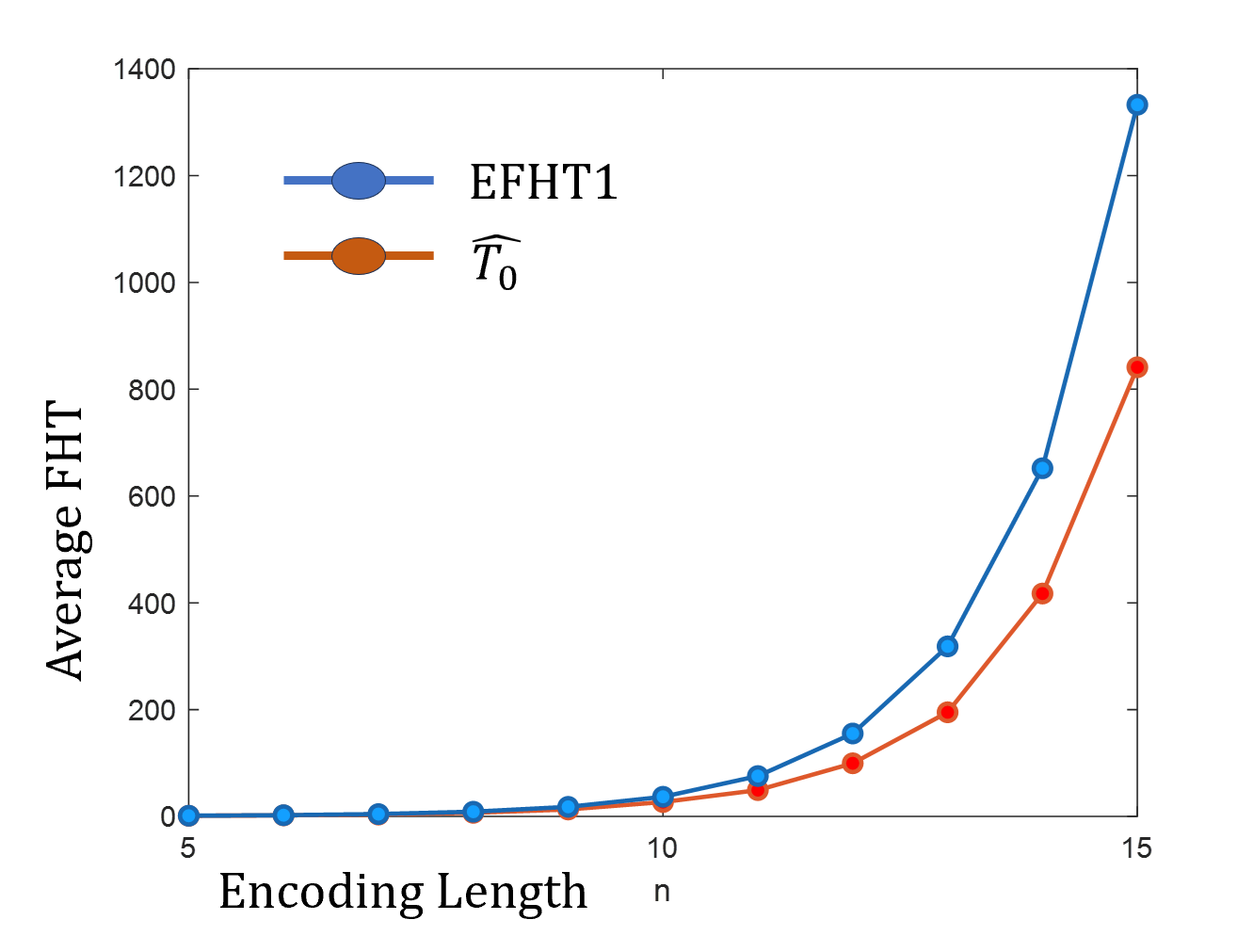}}
	\subfloat[\label{fig:b}]{
		\includegraphics[width=0.33\textwidth]{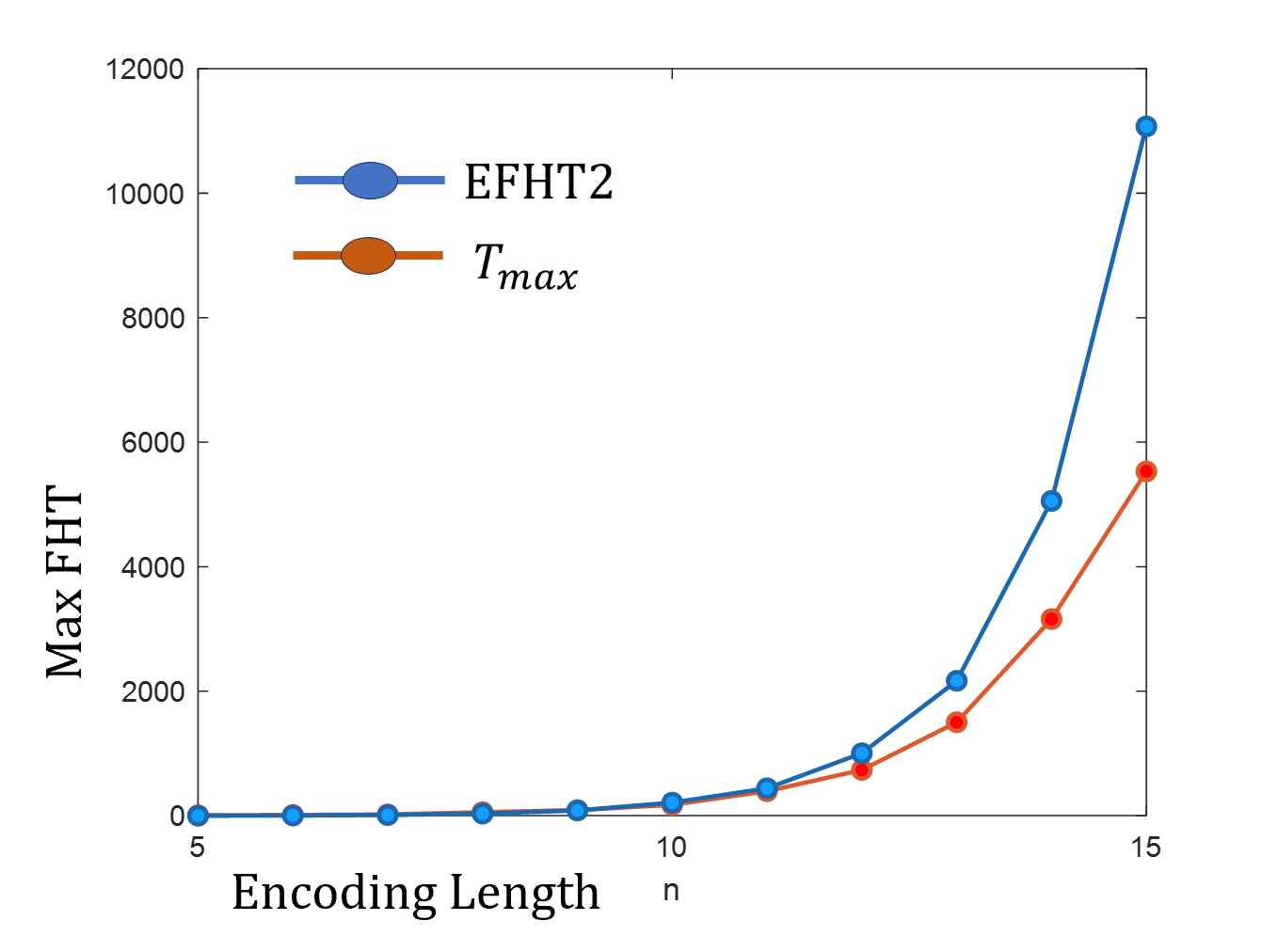}}
	\subfloat[\label{fig:c}]{
		\includegraphics[width=0.33\textwidth]{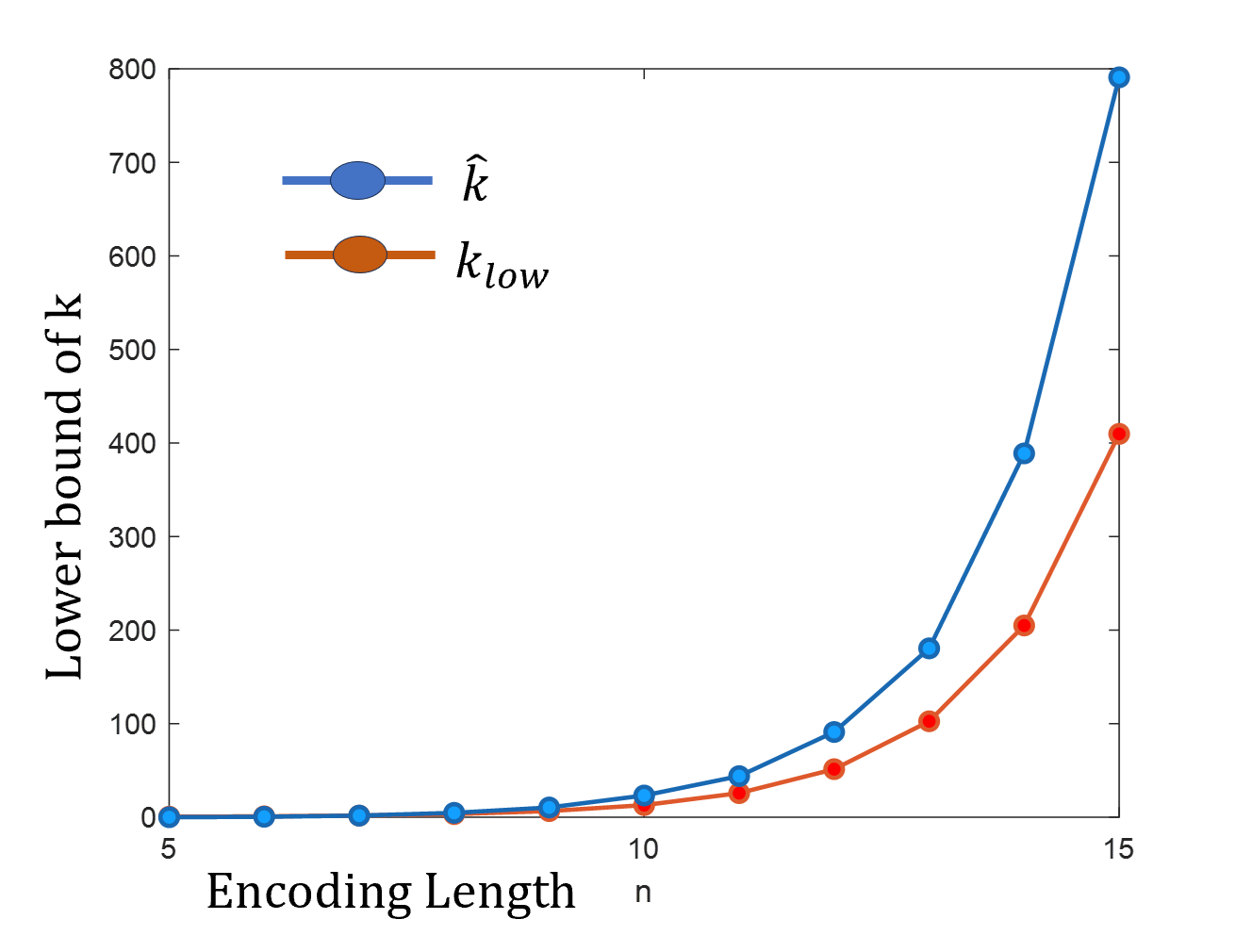}}
	\caption{Results of experiment C. (a) Theoretical average-case upper bounds of EFHT and estimation of actual EFHT. (b) Theoretical worst-case upper bounds of EFHT and actual largest FHT. (c) Estimation of $k_{low}$ and theoretical value of $k_{low}$}
\end{figure*}

\begin{spacing}{1}
	Fig. 4(a) shows the EFHT1 and $\widehat{T_{0}}$ with respect to the encoding length where EFHT1 represents the theoretical upper bound $\sum_{x=1}^{2(n-1)}\frac{1}{x}/\left[(1-e^{\frac{-\lambda\cdot N_{opt}}{2^{n}}})\cdot N_{opt}\right]$ given by Theorem 7. The correlation coefficient between EFHT1 and $T_{0}$ can be obtained as follows:
	\begin{align*}
		r_{3}(EFHT1,T_{0})=0.9999
	\end{align*}
\end{spacing}

\begin{spacing}{1}
	Fig. 4(b) shows the EFHT2 and $T_{max}$ with respect to the encoding length where EFHT2 represents the theoretical upper bound $\widehat{k}Y_{0}/\alpha$ given by Corollary 2. The correlation coefficient between EFHT2 and $T_{max}$ can be obtained as follows:
	\begin{align*}
		r_{3}(EFHT2,T_{max})=0.9935
	\end{align*}
\end{spacing}

\begin{spacing}{1}
	Fig. 4(c) shows the $\widehat{k}$ and $k_{low}$ with respect to the encoding length where $k_{low}$ equals $\beta/\left[(1-e^{\frac{-\lambda\cdot N_{opt}}{2^{n}}})\cdot N_{opt}\right]$ given by Theorem 8. The correlation coefficient between $\widehat{k}$ and $k_{low}$ can be obtained as follows:
	\begin{align*}
		r_{3}(\widehat{k},k_{low})=0.9998
	\end{align*}
\end{spacing}

In Fig. 4, EFHT1$>\widehat{T_{0}}$, EFHT2$>T_{max}$ and $\widehat{k}>k_{low}$. In addition, $r_{3}(EFHT1,T_{0})$, $r_{3}(EFHT2,T_{max})$ and $r_{3}(\widehat{k},k_{low})$ are all greater than 0.91, indicating that the experimental results align with Theorem 7, Theorem 8, and Corollary 2.

\section{Conclusion}
Existing research on running-time analysis of evolutionary combinatorial optimization primarily focuses on simplified problems, and there is a lack of general analysis methods. This paper proposes a multiple-gain model to estimate the average-case and worst-case upper bounds of EFHT for different evolutionary combinatorial optimization instances. The multiple-gain model extends the concept of average gain to provide two general formulas (Theorem 2 and Corollary 2) to estimate the upper bounds of EFHT in both average-case and worst-case scenarios for evolutionary combinatorial optimization. The difference between the multiple-gain model and the average gain model is that the multiple-gain model treats the total gain over multiple iterations in the optimization process as a single gain to overcome the difficulty of the gain lower bound approaching zero in combinatorial optimization. Moreover, experimental data can be utilized in place of theoretical calculations when applying Corollary 2 to estimate the worst-case EFHT.

We utilize the multiple-gain model to estimate the average-case and worst-case upper bounds of EFHT for three different evolutionary combinatorial optimization instances. The estimated results are consistent
with the experimental findings, which demonstrate that the
multiple-gain model is more general than the exsting state-and-the-art methods for evolutionary combinatorial optimization.

Although we analyze different instances of evolutionary combinatorial optimization in this paper, the problems involved are all single-objective optimization problems. The application of the multiple-gain model for the running-time analysis of multi-objective optimization problems will be further investigated.

\bibliography{Multiple-gain}
\bibliographystyle{IEEEtran}

\end{document}